\documentclass[11pt,a4paper]{article}
\usepackage[hyperref]{naaclhlt2019}
\usepackage{times}
\usepackage{latexsym}
\usepackage{latexsym,booktabs,multicol,multirow}
\usepackage{times}
\usepackage{graphicx}
\usepackage[T1,T5]{fontenc}
\usepackage[utf8]{inputenc}
\usepackage{tikz}
\usepackage{amsmath}
\usepackage{xcolor}
\usepackage{collcell}
\usepackage{float}
\usepackage{natbib}
\usepackage[font={small}]{caption}
\usepackage{colortbl,dcolumn}
\usepackage{array}
\usepackage{tablefootnote}

\usepackage{url}

\aclfinalcopy 

\title{Rewarding Smatch: Transition-Based AMR Parsing \\with Reinforcement Learning}

\author{Tahira Naseem$^{\spadesuit}$ ~~~ Abhishek Shah$^{\diamondsuit}$ ~~~ Hui Wan$^{\spadesuit}$ \\ \textbf{Radu Florian}$^{\spadesuit}$~~~\textbf{Salim Roukos}$^{\spadesuit}$ ~~~\textbf{Miguel Ballesteros}$^{\spadesuit}$ \\
  $^{\spadesuit}$IBM Research,
  Yorktown Heights, NY, USA \\
  $^{\diamondsuit}$IBM Watson, New York, NY, USA \\
  {\tt tnaseem, hwan, raduf, roukos@us.ibm.com} \\ {\tt abhishek.shah1, miguel.ballesteros@ibm.com} }
\date{}

\begin{document}
\maketitle
\begin{abstract}



Our work involves enriching the Stack-LSTM transition-based AMR parser \cite{D17-1130} by  augmenting training with Policy Learning and rewarding the Smatch score of sampled graphs. In addition, we also combined several AMR-to-text alignments with an attention mechanism and we supplemented the parser with pre-processed concept identification, named entities and contextualized embeddings. We achieve a highly competitive performance that is comparable to the best published results. We show an in-depth study ablating each of the new components of the parser.

\end{abstract}

\section{Introduction}

Abstract meaning representations (AMRs) \cite{banarescu} are rooted labeled directed acyclic graphs that represent a non inter-sentential abstraction of natural language with broad-coverage semantic representations. AMR parsing thus requires solving several natural language processing tasks; named entity recognition, word sense disambiguation and joint syntactic and semantic role labeling. AMR parsing has acquired a lot of attention  \cite{WangXP15,zhou2,wangxuepradhan,goodman,D18-1198,lyu2018amr,N18-2023,zhang2019amr} in recent years.


We build upon a transition-based parser \cite{D17-1130} that uses Stack-LSTMs \cite{dyer15}. We augment training with self-critical policy learning \cite{rennie2017self} using \emph{sentence-level} Smatch scores \cite{cai2013smatch} as reward. This objective is particularly well suited for AMR parsing, since it overcomes the issues arising from the lack of \emph{token-level} AMR-to-text alignments. In addition, we perform several modifications which are inspired from neural machine translation \cite{bahdanau2014neural} and by the recent trends on contextualized representations \cite{peters2018deep,devlin2018bert}. 

Our contributions are: (1) combinations of different alignment methods: There has been significant research in that direction \cite{flanigan2014discriminative,pourdamghani2014aligning,Chen2015LearningTM,ChuK16,E17-1053,N18-1106,D18-1264}. In this paper, we show that combination of different methods makes a positive impact. We also combine hard alignments with an attention mechanism \cite{bahdanau2014neural}. (2) Preprocessing of named entities and concepts. (3) Incorporating contextualized vectors (with BERT) and compare their effectiveness with detailed ablation experiments. (4) Employing policy gradient training algorithm that uses Smatch as reward. 

\section{Stack-LSTM AMR Parser}
\vspace{-0.1cm}
We use the Stack-LSTM transition based AMR parser of \newcite{D17-1130} (henceforth, we refer to it as BO). BO follows the Stack-LSTM dependency parser by \newcite{dyer15}. This approach allows unbounded lookahead and makes use of greedy inference. BO also learns character-level word representations to capitalize on morphosyntactic regularities \cite{ballesteros15}. BO uses recurrent neural networks to represent the stack data structures that underlie many linear-time parsing algorithms. It follows transition-based parsing algorithms \cite{yamada03,nivre03iwpt,nivre08cl}; words are read from a buffer and they are incrementally combined, in a stack, with a set of actions towards producing the final parse. The input is a sentence and the output is a complete AMR graph without any preprocessing required.\footnote{We refer interested readers to \cite{D17-1130} for details.} We use Dynet \cite{dynet} to implement the parser. In what follows, we present several additions to the original BO model that improved the results.

\subsection{Label Separation}
\vspace{-0.1cm}
BO's actions are enriched with labels that may correspond to AMR nodes or labels that decorate the arcs of the graph. BO reported a total of 478 actions in the 2014 dataset. We tried splitting the prediction in two separate steps, first the action, then the label or concept. This reduces the number of actions to 10 and helps the model to drive the search better.

\subsection{Hard Alignments and Soft Alignments}
\vspace{-0.1cm}
AMR annotations do not provide alignments between the nodes of an AMR graph and the tokens in the corresponding sentence. We need such alignments to generate action sequences with an oracle for training. The parser is then trained to generate these action sequences. The quality of word-to-graph alignments has a direct impact in the accuracy of the parser.

In previous work, both rule-based \cite{flanigan2014discriminative} and machine learning \cite{pourdamghani2014aligning} methods have been used to produce word-to-graph alignments. Once generated, the alignments are often not updated during training \cite{flanigan2016cmu,damonte2016incremental,wang2017getting,foland2017abstract}. More recently, \newcite{lyu2018amr} learn these alignments as latent variables. 

In this work, we combine pre-learned (hard) alignments with an attention mechanism. As shown in section \ref{sec-results}, the combination has a synergistic effect. In the following, we first explain our method for producing hard alignments and then we elaborate on the attention mechanism.  

\paragraph{Hard Alignments Generation:} 

In order to produce word-to-graph alignments, we combine the outputs of the symmetrized Expectation Maximization approach (SEM) of \newcite{pourdamghani2014aligning} with those of the rule-based algorithm (JAMR) of \newcite{flanigan2014discriminative}. \newcite{pourdamghani2014aligning} do not produce alignments for all concepts; for example, named-entity nodes, date-entity nodes and  numerical-quantity nodes are left unaligned. We post-process the output to deterministically align these nodes based on the alignments of its children (if any). We then merge the output with JAMR alignments. Overall, the alignment process involves the following steps:


\begin{enumerate}
    \item Produce initial alignments using SEM\footnote{\url{https://isi.edu/~damghani/papers/Aligner.zip}}.
    \item Fill in the unaligned nodes by upwards percolation of child node alignments.\footnote{When multiple child nodes are aligned, role labels are used to select best node for alignment percolation. Node role labels are preferred in the following order -- :name (in general), \emph{:unit} (for quantities), \emph{:ARG2} (for have-org-role and rate-entities) and then any other labels except \emph{:mod}.}
    \item Use JAMR alignments\footnote{\url{https://github.com/jflanigan/jamr}} for any nodes still unaligned and fill in intermediate nodes again.
\end{enumerate}

\paragraph{Soft Alignments via Attention:}
\label{att}

The parser state is represented by the \textsc{stack}, \textsc{buffer}  and a list with the history of actions (which are encoded as LSTMs, the first two being Stack-LSTMs \cite{dyer15}). This forms the vector  $\mathbf{s}_t$ that represents the state:

\begin{align}
& \mathbf{s}_t = \max \left\{\mathbf{0}, \mathbf{W}[\mathbf{st}_t; \mathbf{b}_t; \mathbf{a}_t] + \mathbf{d}\right\}.
\end{align}


This vector $\mathbf{s}_t$ is used to predict the best action (and concept to add, if applicable) to take, given the state with a softmax. We complement the state with an attention over the input sentence \cite{bahdanau2014neural}. In particular, we use general attention \cite{DBLP:conf/emnlp/LuongPM15}. In order to do so, we add a bidirectional LSTM encoder to the BO parsing model and we run attention over it in each time step. More formally, the attention weights $\alpha_i$ (for position $i$) are calculated based on the actions predicted so far (represented as $a_j$), the encoder representation of the sentence ($h_i$) and a projection weight matrix $W_a$:
\begin{align}
& e_i = a_j^\top W_{a} h_{i} \\ 
& \alpha_i = \frac{\exp(e_i)}{\sum_k \exp(e_k)}.
\end{align}

A vector representation ($c_j$) is computed by a weighted sum of the encoded sentence word representations and the $\alpha$ values.
\begin{align}
c_j = \sum_i \alpha_i \cdot h_i.
\end{align}

Given the sentence representation produced by the attention mechanism ($c_j$), we complement the parser state as follows:
\begin{align}
& g_j = \tanh( W^1_{Dec}  d_j + W^1_{Att}  c_j  ) \\
& u_j = \tanh( g_j + W^2_{Dec}  d_j + W^2_{Att}  c_j  ) \\
&\mathbf{s}_t = \max \left\{\mathbf{0}, \mathbf{W}[\mathbf{st}_t; \mathbf{b}_t; \mathbf{a}_t; \mathbf{u}_j] + \mathbf{d}\right\},
\end{align}
where $d_j$ is the concatenation of the output vector of the LSTM with the history of actions LSTM and the output vector of the LSTM that represents the stack. This new vector $\mathbf{s}_t$ replaces the one described in (1).

Those familiar with neural machine translation will recognize that we are using the concatenation of the output of the LSTMs that represent the stack and the action history as the decoder is used in the standard sequence to sequence with attention model \cite{bahdanau2014neural}.

\subsection{Preprocessed Nodes}

We produce two types of pre-processed nodes: 1) Named Entity labels (NER) and 2) Concept labels (such as want-01, boy etc.). We use NER labels and preprocessed concepts the same way BO and \newcite{dyer15} used part-of-speech tags -- as another vector concatenated to the word representation and learned during training.

\paragraph{Concepts:} AMR representation abstracts away from exact lexical forms. In the case of objects, the concepts are usually represented using the un-inflected base forms; for events, the OntoNotes sense number is attached with the base form (such as \emph{want-01}). We train a linear classifier that uses contextualized BERT embeddings \cite{devlin2018bert} of each word to predict the corresponding concept (which can be none). Each label is predicted in isolation with no regard to the surrounding labels. The tagger is trained using a combination of OntoNotes 5.0 (LDC2013T19) and LDC2017T10 AMR training data.


\paragraph{Named entities:} We extracted named entities from the AMR dataset (there are more than 100 entity types in the AMR language) and we trained a neural network NER model \cite{ni-dinu-florian:2017:Long} to predict NER labels for the AMR parser. In the NER model, the target word and its surrounding words and tags are used as features. We jackknifed (90/10) the training data, to train the AMR parser. The ten jackknifed models got an average NER F1 score of 79.48 on the NER dev set.

\subsection{Contextualized Vectors}

Recent work has shown that the use of pre-trained networks improves the performance of downstream tasks. BO uses pre-trained word embeddings by \newcite{ling2015two} along with learned character embeddings. In this work, we explore the effect of using contextualized word vectors as pre-trained word embeddings. We experiment with recent context based embedding obtained with BERT \cite{devlin2018bert}. 


We use average of last 4 layers of BERT Large model with hidden representations of size 1024. We produce the word representation by mean pooling the representations of word piece tokens obtained using BERT. We only use the contextualized word vectors as input to our model, we do not back-propagate through the BERT layers. 

\begin{table*}[!ht]
\centering
\scalebox{0.68}{
\begin{tabular}{|l|l||c|c|c|c|c|c|c|c|c|}
\hline
Id & Experiment &Smatch & Unlabeled &	No WSD	&Named Entities	&Wikification	&Negations	&Concepts &	Rentrancies	&SRL  \\
\hline
0 & BO (JAMR)  &    65.9 & 71 & 66 & 80 & 0 & 45 & 82 & 46 & 59 \\  
\hline
1 & BO + Label (JAMR)     &      67.0 & 72 &  68 &  81 &  79 &  46 &  82 &  48 &  64 \\
\hline
2& BO + Label     &     68.3 & 73 &  69 &  79 &  78 &  62 &  82 &  51 &  66     \\
3& 2 + POS    &      69.0 &  74 &  70 &  80 &  79 &  62 &  83 & 51 &  67   \\
4&  3 + DEP    & 69.4 & 75 &  70 &  81 &  79 &  65 &  83 &  52 &  67  \\
5&  4 + NER    &      69.8 & 75 &  70 &  83 &  79 &  62 &  83 &  52 &  67  \\
6&  5 + Concepts    & 70.9 & 76 &  71 &  83 &  79 &  66 &  84&  54&  69  \\
7&  6 + BERT    &  72.9 & 78 &  73 &  83 &  78 &  67 &  84 &  58 &   \textbf{72}   \\
\hline
8& 1 + Attention     &    69.8 &  75 &  70 &  80 &  78 &  63 &  83 &  53 &  68   \\
9&  8 + POS    &    70.4 & 75 &  71 &  80 &  79  & 64 &  83 &  53 &  68  \\
10 & 9 + DEP    &      70.7 &  75 &  71 &  80 &  79 &  62 &  83 & 53 &  68 \\
11& 10 + NER    &    70.8 & 76 &  71 &  83 &  79 &  64 & 8 &  53 &  68  \\
12&  11 + Concepts    &      71.8 & 77 &  72 &  82 &  78 &  66 &  84 &  56 &  70   \\
13& 12 + BERT\footnotemark[11]   &      73.1 & 78 &  74 &  82 &  79 &  66 & 84 &  58 &   \textbf{72}  \\
\hline
14&  13 + Smatch     &     73.6 &  78 &  74 &  84 &  79 &  64 &  85&  59 &   \textbf{72}     \\
 \hline
15&  8 + BERT  &      73.4 & 78 &  74 &  83 &  79 &  64 &  84 & 57 &  71   \\
\hline
16&  14 + RL  &      75.5 &  \textbf{80} &  76 &  83 &  80 &  67 &  \textbf{86} &  56 &  \textbf{72}  \\
\hline
\hline
&\newcite{zhang2019amr}    &      \textbf{76.3}  & 79 & \textbf{77} & 78 & \textbf{86} & \textbf{75} & 85 & \textbf{60} & 70 \\
&\newcite{lyu2018amr}    &    74.4 & 77 & 76 & \textbf{86} & 76 & 58 & \textbf{86} & 52 & 70 \\
&\newcite{NoordB17a} &      71.0 & 74 & 72 & 79 & 65 & 62 & 82 & 52 & 66 \\
&\newcite{D18-1198}    &      69.8 &  74 & 72 & 78 & 71 & 57 & 84 & 49 & 64   \\
\hline

\end{tabular}

}
\caption{Results, including comparison with the best systems, in the LDC2017T10 test set (aka AMR 2.0). Results highlighted in bold are the best in each metric. BO is \cite{D17-1130} (which did not produce wikification). (JAMR) means that the model uses JAMR alignments, the rest use our alignments. Metrics by \newcite{cai2013smatch} and \newcite{damonte2016incremental}.}
\label{results}

\end{table*}

\subsection{Wikification}

Given that BO does not produce Wikipedia nodes during prediction, we pre-process the AMR data removing all Wikipedia nodes. In order to produce Wikipedia entries in our AMR graphs, we run a wikification approach as post-processing. We combine the approach of \newcite{lyu2018amr} with the entity linking technique of \newcite{sil2018neural}. 

First, we produce a dictionary of Wikipedia links for all the named entity nodes that appear with \emph{:wiki} label in the training data. If a node appears with multiple Wikipedia links, the most frequent one is added to the dictionary. Separately, we also process the target sentence using the entity linking system of \newcite{sil2018neural}. This system identifies the entities as well as links them. 

During post processing, every node with \emph{:name} label is looked up in the dictionary and if found, is assigned the corresponding Wikipedia link. This is very similar to the approach of \newcite{lyu2018amr}. If the node is not found in the dictionary, and the system of \newcite{sil2018neural} produces a Wikipedia link, we use that link.

\subsection{Smatch Weighting}

The upper bound for BO's oracle is only 93.3 F1 for the entire development set. We observed that the oracle produces a score close to perfect for most sentences, yet it loses some points in others. During training, we have the gold AMR graph available for every sentence. We compare it to the oracle graph and use the Smatch score as a weight for the training example. This is a way to down-weight the examples whose oracle actions sequence is incomplete or erroneous. This modification resulted in moderate gains (see row 14 in Table \ref{results}) and also lead to the training with exploration experiments described below.

\section{Reinforcement Learning}

BO relies on the oracle action sequences. The training objective is to maximize the likelihood of oracle actions. This strategy has two drawbacks. First, inaccurate/incomplete alignments between the tokens and the graph nodes.(As mentioned above, the oracle upper bound is only 93.3 F1. With the enhanced alignments, BO reported 89.5 F1 in the LDC2014 development set). Second, even for the perfectly aligned sentences, the oracle action sequence is not the only or the best action sequence that can lead to the gold graph; there could be shorter sequences that are easier to learn. Therefore, strictly binding the training objective to the oracle action sequences can lead to sub-optimal performance, as evidenced in \cite{daume05,daume09,goldberg12coling,goldberg2013training,D16-1211} among others.  

To circumvent these issues, we resort to a Reinforcement Learning (RL) objective where the Smatch score of the predicted graph for a given sentence is used as reward. This alleviates the strong dependency on hard alignment and leaves room to training with exploration of the action space. This line of work is also motivated by \newcite{goodman}, who used imitation learning to build AMR parses from dependency trees.

We use the self-critical policy gradient training algorithm by \newcite{rennie2017self} which is a special case of the REINFORCE algorithm of \newcite{williams1992simple} with a baseline. This method allows the use of an external evaluation measure as reward \cite{paulus2017deep}. In particular, we want to maximize the expected Smatch reward,

\begin{align}
L_{RL}=E_{g^s \sim p_{\theta}}[r(g^s)] \label{rlobj}
\end{align}

where $p_{\theta}$ is the policy specified by the parser parameters $\theta$ and $g^s$ is a graph sampled from $p_{\theta}$. The gradient of this objective can be approximated using a single sample from $p_{\theta}$. For each sentence, we produce two graphs using the current model parameters. A greedy best graph $\hat{g}$ and a graph $g^s$ produced by sampling from action space. The gradient of \ref{rlobj} is approximated as in \cite{rennie2017self},

\begin{align}
    \nabla_{\theta} L_{RL}=(r(g^s)-r(\hat{g})) \nabla_{\theta} \log(p_{\theta}(g^s)) 
\end{align}

where $r(g)$ is the Smatch score of graph $g$ with respect to the ground truth. The Smatch of the greedy graph $r(\hat{g})$ serves as a baseline that can reduce the variance in the gradient estimate \cite{williams1992simple}. 

With $\epsilon$ probability, we flatten the sampling distribution by calculating the square root of the probabilities. In our experiments, $\epsilon$ is set to $0.05$. We first train our full model with the  maximum-likelihood objective of BO that achieves an F-score 72.8 without beam search when evaluated in the development set. The RL training is then initialized with the parameters of this trained model. For RL training, we use a batch-size of 40. 

\section{Experiments and Results} \label{sec-results}

We start by reimplementing BO\footnote{BO reported results on the 2014 dataset.} and we train models with the most recent dataset (LDC2017T10)\footnote{LDC2016E25 and LDC2017T10 contain the same AMR annotations as of March 2016. LDC2017T10 is the general release while LDC2016E25 was released for Semeval 2016 participants \cite{may2016semeval}.}. We include label separation in our reimplementation (Experiments 1..16) which separates the prediction of actions and labels in two different softmax layers. All our experiments use beam 10 for decoding and they are the best (when evaluated in the development set) of 5 different random seeds. Word, input and hidden representations have 100 dimensions (with BERT, input dimensions are 1024), action and label embeddings are of size 20. Our results are presented in Table \ref{results}.

We achieve the best results ever reported in some of the metrics. Unlabeled Smatch (16) by 1 point and SRL by 2 points. These two metrics represent the structure and semantic parsing task. For all the remaining metrics, our parser consistently achieves the second best results. Also, our best single model (16) achieves more than 9 Smatch points on top of BO (0). \newcite{D18-1198}'s parser is a reimplementation of BO with a refined search space (which we did not attempt) and we beat their performance by ~5 points. 

The hard alignments proposed in this paper present a clear advantage over the JAMR alignments. BO ignores nodes that are not aligned to tokens in the sentence, and it benefits from a more recall oriented alignment method. Adding attention on top of that adds a point, while preprocessing named entities improve the NER metric. Adding concepts preprocessed with our BERT based tagger adds more than a point. Smatch weighting lead to half a point on top of (14). 

BERT contextualized vectors provide more than a point on top of the best model with traditional word embeddings (without attention, the difference is of 2 points). Combining BERT with a model that only sees words (15), we achieve the best results surpassed only by models that also use contextualized vectors and reinforcement learning objective, However, we added Smatch weighting (14) and Reinforcement Learning (16) on top of 13. This was decided based on development data results, where 13 performed better than the BERT only model (15) by about a point. 

Finally, training with exploration via reinforcement learning gives further gains of about 2 points and achieves one of the best results ever reported on the task and state of the art in some of the metrics.


\section{Conclusions}

We report modifications in a competitive AMR parser achieving one of the best results in the task. Our main contribution augments training with Policy Learning by priming samples that are more suitable for the evaluation metric. We perform an in-depth ablation experiment that shows the impact of each of our contributions. Our unlabeled Smatch score (achieving the best graph structure) suggests that a new strategy to predict labels may reach even higher numbers.

\section*{Acknowledgments}
We thank Ram\'on Astudillo, Avi Sil, Young-Suk Lee, Vittorio Castelli and Todd Ward for useful comments and support.

\bibliography{naaclhlt2019,main}
\bibliographystyle{acl_natbib}
\end{document}